\documentclass[]{article}

\usepackage{graphicx}

\usepackage{naacl2001}
\usepackage{qtree}

\title{Modeling informational novelty in a conversational system with a hybrid statistical and grammar-based approach to surface natural language generation}
\author{Adwait Ratnaparkhi\\IBM TJ Watson Research Center\\PO Box 218\\Yorktown Heights, NY 10598\\aratnapa@us.ibm.com}

\begin{document}

\maketitle

\begin{abstract}
We present a hybrid statistical and grammar-based system for surface
natural language generation (NLG) that uses grammar rules, conditions on
using those grammar rules, and corpus statistics to determine the word
order.  We also describe how this surface NLG module is implemented in
a prototype conversational system, and how it attempts to
model informational novelty by varying the word order.  
Using a combination of rules and statistical
information, the conversational system expresses the novel information
differently than the given information, based on
the run-time dialog state.  We also discuss our plans for evaluating
the generation strategy.  
\end{abstract}

\section{Introduction}

We present a module for surface natural language generation (NLG) that is
capable of dynamically re-ordering words based on information
in the run-time dialog state of a conversational system in the air travel domain.
For our purposes, we make the distinction
that deep NLG is the process of deciding what information to convey, 
whereas surface NLG is the process of rendering that information in natural language. 
The surface NLG module is used by the conversational system to express the
 new information differently than old information, similar to how people
might express informational novelty in a human-human conversation. 
The eventual goal of this experiment is to test if strategically 
re-ordering the words in the response of a conversational system
can address a widely held criticism, namely, 
that such systems ``don't sound human.''

\section{Previous Approaches}

The most popular technique for surface NLG is templates. A template for
describing a flight noun phrase in the air travel domain might be
``flight departing from \$city-fr at \$time-dep and arriving in \$city-to
at \$time-arr''
where the words starting with ``\$'' are actually variables
---representing the departure city, and departure time, the arrival
city, and the arrival time, respectively--- whose values will be
extracted from the environment in which the template is used.
This approach requires the programmer to write a different template for every possible word ordering, 
and may be impractical for domains in which many word orderings are necessary.

There are more sophisticated surface generation packages,
such as FUF/SURGE \cite{surge96}, KPML \cite{kpml:tech}, MUMBLE \cite{mumble}, 
and RealPro \cite{realpro}, which produce natural language text 
from abstract semantic representations. These packages use many rules written 
by linguistic experts to map the input representations to textual output.

In order to partially automate the process of mapping input representations to
textual output, several researchers have recently investigated the use of statistics
in generation.
Our approach is in the same spirit as other recent work, such as
such as \cite{langkilde98}, \cite{ratnaparkhi:naacl2000}, and \cite{bangalorerambow2000}, 
in which statistics from a corpus have been used to disambiguate or rank candidates for surface generation. 
We compare our approach with these approaches in section~\ref{comparison}.

\section{Our approach: A Hybrid Grammar and Statistical surface NLG system}

The motivation for our approach is the desire for a surface generation framework that allows dynamic
word re-ordering in the context of a conversational system. 
People naturally re-order words in a conversation in a way that maximizes their communicative power, 
and we hope to duplicate this behavior in an automated conversational system. 
We describe our surface generation framework, the linguistic behavior it tries to duplicate, and its implementation 
in an air travel conversational system.

The central idea of our surface NLG module is that given  a dependency-like grammar, 
it generates many word sequences that are consistent with the grammar rules and rule conditions,
and uses corpus statistics to find the word sequence that most resembles the real utterances of people.
In our framework, the input to the surface NLG module consists of 
\begin{itemize}
\item a {\em mandatory} set of attribute-value pairs, $A_1$
\item an {\em optional} set of attribute-value pairs, $A_2$
\end{itemize}
In the air travel domain, an example attribute might be ``\$city-fr'', denoting the departure city, 
while an example value might be ``New York''. (Attributes are denoted here by the ``\$'' prefix.)
The surface NLG is required to express every attribute in the set $A_1$, but does not need to 
express anything from the set $A_2$. 
In practice, $A_1$ carries the intended meaning of the utterance to be generated, 
while $A_2$ carries miscellaneous information
that is accessible to the NLG module.
The sets $A_1$ and $A_2$ are extracted from the dialog state of the conversational system, after the dialog manager has
determined that it needs to speak an utterance to the user, and after the deep generation module has determined the content of $A_1$ and $A_2$.
We assume that any relevant fact about the discourse history exists in the dialog state, and can be encoded as an attribute-value pair in either $A_1$ or $A_2$. 
We make the assumption that attribute-value pairs are sufficient to describe the meaning of the utterance
we intend to generate. This assumption  is reasonable for small domains like air travel.

\subsection{Grammar}

In our approach, grammar rules define the possible dependency trees the NLG module may generate in the context
of the current dialog state.
Any dependency tree generated by our grammar can be converted to a flat word sequence by a deterministic procedure.

A grammar rule in this system specifies a relationship between a parent, and one more children, using the following structure:
\begin{description}
\item[Parent:] This is the parent, and is usually the linguistic ``head'' of the phrase.
\item[Direction:] is either - (left) or + (right), and indicates the intended word order of the children relative to the parent.
\item[Children:] One or more words that are children to the parent
\item[Condition:] A code fragment that evaluates to either true or false in the current state of the dialog system ($A_1$, $A_2$)
\end{description}
The order of the children specified in a single rule is fixed; it is merely the order written by the programmer. 
However,  there is no ordering constraint between the children of {\em different} rules with the same head.
For example, take the following grammar (in which the {\bf Condition} section has been omitted for clarity), 

\begin{tabular}{ccc}
Parent&Direction&Children\\ \hline
a&+&b\\
a&+&c d\\
c&-&f\\
\end{tabular}

This grammar will allow the following dependency trees,

\Tree [.a b+ ]
\Tree [.a c+ d+ ]

\noindent each of whom represent the application of one grammar rule for the head ``a''.
The + sign denotes a right child, whereas the - sign denotes a left child (regardless of whether the child is visually typeset to the left or right of its parent). 
The following trees reflect the use of the first two rules, in both possible orderings

\Tree [.a b+ c+ d+ ]

\Tree [.a c+ d+ b+ ]

The siblings ``c'' and ``d'' cannot be re-ordered or broken up 
with respect to each other, since they were specified in a single rule. So the tree

\Tree [.a c+ b+ d+ ]

\noindent is disallowed.
The children can themselves be recursively expanded, so the tree

\Tree [.a b+ [.c+ f- ]  d+ ]

\noindent is allowed. The dependency trees can be converted to word sequences (i.e., linearized), by recursively traversing the left children, the parent, and the right children. The word sequence corresponding to the tree above is ``a b f c d''. It is possible for different dependency trees to yield the same word sequence. 

\begin{table*}
\begin{tabular}{|cccp{4in}|} \hline
Parent&Direction&Children&Condition\\ \hline
flights&+&from New York&{ value of departure-city in dialog state is ``New York'' }\\ \hline
\end{tabular}
\caption{Sample rule to describe departure city}
\label{samplerule1}
\end{table*}

\begin{table*}
\begin{tabular}{|cccp{4in}|} \hline
Parent&Direction&Children&Condition\\ \hline
flights&+&from \$city-fr&{ the \$city-fr attribute exists in the dialog state }\\ \hline
\end{tabular}
\caption{Sample rule with attribute to describe departure city}
\label{samplerule2}
\end{table*}

The {\bf Condition} section of the rule specifies an arbitrary code fragment that is  
evaluated in the context of the attribute-value sets $A_1, A_2$, which are derived from 
the current dialog state. The rule is used only if the code fragment evaluates to true. 
A rule condition associates an element of meaning with its realization as a phrase in natural language.
For example, in the air travel domain, Table~\ref{samplerule1} lists a grammar rule, with a condition in pseudo-code,  that might be used to describe a departure city.
This rule would allow ``flights from New York'' if the departure city, as specified in the dialog state, is ``New York''.

The rule condition can be more complex if necessary; 
in our implementation, the rule condition is an arbitrary fragment of code in the language Tcl. 
Also, the rules in our implementation are slightly more abstract in that they may 
contain attributes in addition to words. In this case the attribute
will be instantiated with a value of interest at some later point.
Table~\ref{samplerule2} shows a rule for describing departure cities that uses attributes.
With attributes allowable as children, 
the output of the NLG module is essentially just a template. 
The difference between our system and the template method for NLG is that
the programmer need only specify template {\em fragments} in the form of parent/children relationships, 
instead of the entire template.

Lists with conjunctions are linguistic phenonema that need to be generated frequently in the air travel domain.
For example, a list with one item $a_1$ is realized  simply as ``$a_1$'', but two items are realized as 
``$a_1$ and $a_2$'', while $n > 2$ items are realized as ``$a_1$, $\dots$, $a_{n-1}$,  and $a_n$''.
We found it easier to properly generate the conjunction and commas 
with a built-in construct, as opposed a programmer-supplied grammar rule. 
We define the constructs ``\&'' and ``$|$'' to denote that the children of a rule must be 
generated with conjunctions (``and'' and ``or'', respectively) and commas.
For example, the grammar (with the conditions omitted):

\begin{tabular}{ccc}
Parent&Direction&Children\\ \hline
a&+\&&b\\
a&+\&&c\\
a&+\&&d\\
\end{tabular}

\noindent will generate (among others) the word sequences
\begin{itemize}
\item a b
\item a b and c
\item a b , c , and d
\end{itemize}
Of course, the words ``and'' and ``or'', are dependent on the language and perhaps even the genre of the language. 
The comma notation is dependent on the application of interest. It is critical to have the commas placed properly if 
the generated text will eventually be synthesized into speech; the speech synthesizer relies on commas to generate
the appropriate pauses in the speech output.

The system currently has no facility that is specially designed to handle the generation of morphological variants.
For example, in the air travel domain, 
the word ``flight'' should be realized as ``flights'', if the 
number of flights is greater than 1. Similarly, the verb ``arrive'' must agree with ``flight'' in the phrase ``flight that arrives'' versus ``flights that arrive''. We instead use a generic token re-write facility, in which the programmer, 
for example, can tell the system to  re-write 
the word ``flight'' as ``flights'' based on information in $A_1, A_2$. 
This facility can be flexible since different uses of the same word in the grammar can be represented by distinct 
tokens, (e.g. flight-subj, flight-obj) which are later realized into their morphologically correct spellings. 
Our system is not meant to be a general purpose generator, and 
in the future, we plan to extend our system to better handle the generation of morphological variants.


\subsubsection{Assigning scores to trees}

We assign scores to a dependency tree $t$ by first converting it to a word sequence $w_1 \dots w_n$, and by using an interpolated $n$-gram language model on the word sequence:
\begin{eqnarray*}
P(w_1 \dots w_n ) &=& \prod_{i=1}^n P(w_i | w_{i-1} \dots w_1 )\\
P(w_i | w_{i-1} \dots w_1 )&=& \sum_{j=1}^4 \lambda_j P_j(w_i | w_{i-1} \dots w_1 )
\end{eqnarray*}
The probability models $P_j$ are computed from statistics derived from roughly 8000 utterances in the air travel domain.
The probability model $P_1$, $P_2$, and $P_3$ are  derived from trigram, bigram, and unigram statistics, while $P_4$ is the uniform model. The $\lambda_j$ are set heuristically such that $\lambda_j \geq 0$ and 
$\sum_{j=1}^4 \lambda_j = 1$.

\subsection{Searching for the best dependency tree}

The goal of the system is to find the highest-scoring dependency tree that is consistent with 
the grammar. The strategy is, given an existing tree $t$,  to enumerate all the ways of creating 
new dependency trees $t_1 \dots t_n$ that are consistent with the grammar, 
and to only keep the top $N$ scoring trees for consideration in the next search iteration. 
The search terminates when $N$ $A$-completed\footnote{A mnemonic for ``attribute-completed''} 
trees are found, where an $A$-completed tree mentions all of 
the attributes in the mandatory attribute-value set $A_1$ {\em exactly once}. 
We justify the restriction to $A$-completed trees because trees that have omitted one or more
attributes are clearly not expressing the meaning of $A_1$, while we view trees that mention 
the same attribute more than once as containing redundant information.
The highest scoring $A$-completed tree is the answer returned by the NLG module.
This search strategy is heuristic in nature; it is not guaranteed to find the highest-scoring tree.

\begin{figure}
\includegraphics[scale=.5,height=3in]{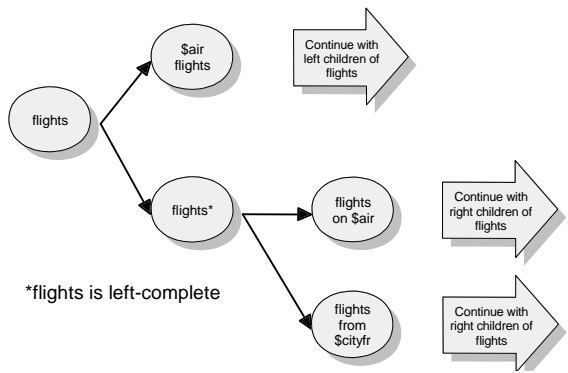}
\caption{Pictorial depiction of search algorithm}
\label{searchfigure}
\end{figure}

On each search iteration, the system finds the top scoring $N$ trees $t_1 \dots t_n$ that are currently under consideration, 
and attempts to create a new set of trees by applying the following algorithm to each tree $t$ in the set. 
\begin{itemize}
\item Check to see if $t$ is $A$-complete. If so, remove it from consideration. If $N$ trees are $A$-completed, terminate the search.
\item If $t$ is not $A$-complete, 
the system determines the {\em active parent}, by starting at the root of $t$, and recursively checking the
left children, the right children, and then the parent itself, for the first tree node that is not {\em completed}. 
A tree node is completed if 
\begin{itemize}
\item it is {\em left-complete}, meaning that all of its left children have been generated, and
\item it is {\em right-complete}, meaning that all of its right children have been generated
\end{itemize}
\item If no active parent is found, $t$ is discarded, since we cannot apply more rules to make $t$ $A$-complete.
\item If $p$ is the active parent, the system decides to work in the left direction if $p$ is not left-complete, otherwise it works in the right direction.
\item Either
\begin{itemize}
\item Apply a rule: once the direction is settled, the system applies a rule $r$ in the grammar if
\begin{itemize}
\item the parent specified in $r$ is equal to the active parent $p$
\item the condition of $r$ evaluates to true  
\item $r$ has not been previously used to generate children for the parent $p$
\item the attributes mentioned in the children have not been mentioned elsewhere in the tree
\end{itemize}
\item If the rule can be applied, add the children in the rule to the active parent. Add from right-to-left if we are adding left children, add from left-to-right if we are adding right children.
\item Use the new tree $t'$ for consideration in the next search iteration
\end{itemize}
\item or mark the tree 
\begin{itemize}
\item left-complete, if we were adding in the left direction
\item right-complete, if we were adding in the right direction
\item use the new tree $t''$ for consideration in the next search iteration
\end{itemize}
\end{itemize}

The point of the search algorithm is that it explores many possible word sequences, while
requiring the programmer to only specify template fragments in the form of dependency tree parent/children relationships. 
The programmer can specify several ways to express any given attribute; the search guarantees that any
attribute given in $A_1$ will be mentioned only once in the generated utterance.
Intuitively, the system takes the fragments of natural language 
given by the programmer, explores many ways of ``pasting'' them together such that they respect the grammar, and 
returns the ``best'' way with respect to the scoring function.
Figure~\ref{searchfigure} gives a pictorial depiction of the search procedure looking for ways to express the attributes \$city-fr and \$air, which represent the departure city and air carrier, respectively.
The attributes are instantiated with their corresponding values after the search has found the best candidate
for surface generation.

\section{Using word order to express informational novelty}

\label{newoldsection}
It has been long argued that utterances  have an {\em information structure}, such that one 
part refers to pre-existing information in the discourse, while the other 
part refers to information that is newly introduced into the discourse.
There are several existing dichotomies,  which capture the same general idea but differ
in their details, such as {\em theme vs rheme}, {\em topic vs. comment}, {\em presupposed vs. focus}. 
See \cite{prevost:phd} for a summary of different information structure schemes.

We want to model informational novelty, which correlates roughly with the theme vs. rheme distinction, 
so that old information (theme) is expressed differently than the new information (rheme).
Furthermore, at this time, we wish to do it without modifying the pitch, only with word ordering. 
\cite{steedman:icslp96} gives a more fine-grained information structure, and points out that sub-elements of the
theme can also contain new information,  and are often  emphasized with pitch. 
However, for our purposes, we choose to model the more simplistic structure of new versus old information, 
since this is the only distinction we can reliably make in our current dialog system.
 
In  a spoken conversational system, it is usually necessary to confirm to the user what was spoken and understood by the computer in the last turn. 
This way, the user can ascertain if system's speech recognition module and natural language understanding module  are working correctly, and can repeat any information that the computer misunderstood. 
Another reason for  confirmation messages is to remind the user 
of information that was understood several turns ago. 
Ideally, we should analyze a sample of text in the domain of interest (air travel) and annotate
how the confirmation information (new vs. old) is expressed. However, most confirmations in human-human 
conversations do not contain both new and old information, and 
happen in a manner that is not easily reproducible with a speech-to-speech conversational system, as shown
in Table~\ref{confirm1}. In this type of dialog, the user interrupts the travel agent in order to confirm 
what has been recently spoken, the confirmation is done with an ``mm hmm'' sound. 
We have noticed that some confirmations, do contain both old and new information, as shown in Table~\ref{confirm2}.
In this case, the old information (``Buffalo to Chicago'') is spoken  for confirmational purposes. 

Many of the human-human dialogs in the air travel domain that do contain old and new information are expressed in a way such 
that the old information precedes the new information. At this time, we are still accumulating quantitative evidence 
in a corpus of transcribed human-human dialogs in the air travel domain
to make this claim more precise. Furthermore, in English, it has been long noted that
there is a tendency to specify old information before new information (e.g., see \cite{sornicola99:topicfocus} 
for a review of many studies).  

Note that word order is not the only indication of novelty! 
It is clear that other factors, such as pitch and loudness, also convey novelty, even when the word
order is fixed. We assume that pitch and loudness are roughly constant, and that we can model
novelty by only varying the word order. Also, using only word order to model novelty has the advantage
that it leaves open the possibility of using our NLG technique for non-spoken text, e.g., an interactive web page.

\begin{table}
\begin{tabular}{lp{2.5in}}
Agent:&we have you returning on the seventeenth of September on US Air flight five zero seven\\
User:&mm hmm...\\
Agent:&out of Syracuse at seven fifty a.m. into Pittsburgh at nine oh five a.m.\\
User:&mm hmm...\\
\end{tabular}
\caption{Example of confirmation only in human-human dialog}
\label{confirm1}
\end{table}

\begin{table}
\begin{tabular}{lp{2.5in}}
User:&What was the Buffalo to Chicago flight ?\\
Agent:&ah Buffalo to Chicago is three ninety three \\
\end{tabular}
\caption{Example of agent confirming old information and introducing new information in human-human dialog}
\label{confirm2}
\end{table}

\section{Application: Modeling informational novelty in a conversational system}

The hybrid surface NLG module has been integrated into a
telephony conversational system for air travel reservations, developed for the 
{\sc DARPA Communicator} effort, and described in \cite{axelrod:nlg}. 
Most system utterances are generated using an existing template-based approach, while 
a certain class of utterances are generated with the NLG system described in this paper.

The conversational system first collects information from the user, and then consults a flight
database to find flights that match the user's constraints.
If one flight was found, it asks the user to confirm it. 
If no flights were found, it asks the user to relax some of the constraints, whereas if many flights were
found, it prompts the user to further constrain the flight list. In the case where either many or no flights were found, 
the first utterance given by the system is called a {\em summary sentence}, whose purpose is to give a 
one-sentence summary of the results from the flight database.
In the current system, the summary sentence is generated using templates, where some template fragments are 
``optional'', so that they are printed only in the presence of certain attributes. 
In the existing approach, the generation of certain words is optional, but the order in 
which they are presented is fixed.

In our new approach, the word order in the summary sentence depends on three sources of information
\begin{description}
\item[Grammar:] This is the dependency grammar specified by the programmer. Approximately 50 rules were needed to 
generate the possible summary sentences.
\item[Statistics:] These are derived from a corpus of roughly 8000 utterances in the air travel domain, and are
used by the NLG module's scoring function.
\item[Attribute Novelty:] 
Each attribute in set of mandatory attributes $A_1$ is marked as either {\em old} or {\em new}. 
For our purposes, {\em new} attributes are those which were given to the system in the last user turn. 
Anything not marked new is assumed to be {\em old}. 
\end{description}
The attributes are marked as old or new by the deep generation component, by using information 
in the dialog history and some heuristics.

We use the surface NLG module to express the new attributes differently than  the old attributes in the summary sentence.
The surface NLG system allows us to detect the novelty of an attribute in any of the
rule conditions of the grammar, e.g., with a function call that takes the name of an attribute
and returns either true or false.
Therefore,  we can have two kinds of rules for every attribute: one rule to express it when it is new),
and another rule to express it when it is old. 
The grammar for the summary sentence is written in a way to produce sentences 
having  general structure shown in Figure~\ref{summarystructure}.
The [old information] is the area in the sentence in which the old information
will be expressed, while the  [new information] is the area in the sentence 
in which the new information will be expressed.
In the case where there is more than one old and one new attribute, novelty alone does not determine
the word order; it merely tells the NLG module the area in the sentence that will contain the  phrase expressing
the attribute.
The complete word order is attained with the scoring function (applied in the search procedure), 
which  ranks the possible different orderings that are consistent with the grammar structure.

\begin{figure*}
\begin{center}
\fbox{
{\small There are N flights [old information] that [new information]}
}
\end{center}
\caption{General structure for flight database summary sentence in air travel conversational system}
\label{summarystructure}
\end{figure*}

\begin{table*}
\begin{tabular}{|cccp{4in}|} \hline
Parent&Direction&Children&Condition\\ \hline
flights&+&from \$city-fr&{ the \$city-fr attribute exists in $A_1$ , \$city-fr is old  } \\ \hline
leave&+&from \$city-fr&{ the \$city-fr attribute exists in $A_1$, \$city-fr is new  } \\ \hline
flights&+&that&{ if at least one attribute is marked new } \\ \hline
that&+&leave&{ if \$city-fr, \$date-dep, or \$time-dep are new  } \\ \hline
\end{tabular}
\caption{Sample rules to describe departure city as either old or new information}
\label{oldandnewrules}
\end{table*}

Table~\ref{oldandnewrules} shows how several grammar rules can be used to control the placement of old and new information.
Table~\ref{oldsystem} contains a dialog with the existing conversational air travel system.
Table~\ref{newsystem} contains the dialog, with the same user input, 
but with the system using the NLG module discussed in this paper.
Approximately 50 grammar rules were used to generate the summary sentence in our prototype conversational system.

In our approach, the definition of novelty is rather peculiar. 
We mark an attribute as {\em new} if it has been introduced by the user in 
the previous turn, and mark it as {\em old} otherwise. 
Using this definition, all of the attributes in the summary sentence are ``old'' in the sense that they 
have already been introduced into the dialog by the user; the labels {\em new} and {\em old} only distinguish
the time at which the user introduced the attributes.
Truly novel information, i.e., information from the flight database that has not previously been heard by the user or 
said by the system, is not expressed in the summary sentence. 
Unfortunately, in our current approach, we confine the use of the hybrid NLG system to the summary sentence. 
In future work, we plan to extend it to other kinds of sentences in the system's response in the hopes of 
modeling the ``truly novel'' information.

\begin{table}
\begin{tabular}{cp{2.5in}}
User&i want to fly from new york to pittsburgh tomorrow at ten\\
Sys&do you want to leave about ten A M or at ten P M? \\
User&ten A M\\
Sys&\textbf{i see a couple of flights from new-york to pittsburgh which depart about ten A M on september nineteenth.} the first flight is on T W A.. It departs at nine forty A M and arrives at eleven sixteen A M... the second flight is on U S air. It departs at ten twenty five A M and arrives at eleven forty one A M..\\ 
User&i want delta\\
Sys&\textbf{i can't find any flights from new-york to pittsburgh on delta which depart about ten A M on september nineteenth.} you may want to try changing the airline to T W A or U S air.\\
\end{tabular}
\caption{Dialog with existing conversational system. Summary sentence is in bold face.}
\label{oldsystem}
\end{table}

\begin{table}
\begin{tabular}{cp{2.5in}}
User&i want to fly from new york to pittsburgh tomorrow at ten\\
Sys&do you want to leave about ten A M or at ten P M? \\
User&ten A M\\
Sys&\textbf{there are several flights from new-york to pittsburgh on september nineteenth that {\em leave around ten A M.}} the first flight is on T W A.. It departs at nine forty A M and arrives at eleven sixteen A M... the second flight is on U S air. It departs at ten twenty five A M and arrives at eleven forty one A M.. \\
User&i want delta\\
Sys&\textbf{there are no flights from new-york to pittsburgh on september nineteenth around ten A M that {\em are served by delta}}. you may want to try changing the airline to T W A or U S air. \\
\end{tabular}
\caption{Dialog with conversational system using surface NLG module for summary sentence. Summary sentence is in bold face. New attributes are italicized.}
\label{newsystem}
\end{table}

\section{Evaluation}

Evaluating our surface NLG system in isolation is difficult, because its output is dependent on information
in the dialog state of a conversation. 
We propose to evaluate it in the context of an entire conversational system, 
by recording the time it takes for a user to complete certain tasks using the 
existing template generation, and comparing it with how long it takes users to complete the same tasks using the
surface NLG module described in this paper. 
Such  evaluations are already underway in the {\sc DARPA Communicator} effort \cite{communicator:eval}.
We believe that strategically setting the word order will reduce the amount of attention and mental effort necessary 
from the user in order to successfully use a conversational system.  We hope that this reduction in mental effort 
will allow users to have higher success rates with the system, faster completion times, and eventually, the ability
to multitask, i.e., to ability to use the system while engaged in some other secondary activity. 

\section{Comparison with other work}
\label{comparison}

Our work is similar to  \cite{langkilde98}, \cite{bangalorerambow2000}, and \cite{ratnaparkhi:naacl2000} 
in that we use statistical information to select between 
multiple candidates for surface generation. \cite{langkilde98} use statistics to 
select the best generation candidate from a  word lattice generated from a grammar, while \cite{bangalorerambow2000} 
use statistics to select the word order of an underspecified dependency tree generated from a grammar. 
\cite{ratnaparkhi:naacl2000} uses statistics to rank candidates given by a grammar induced from 
a dependency tree annotated corpus. 

Our approach differs from previous approaches in that it is specifically
directed towards modeling informational novelty in a conversational system.
In our system, the programmer can impose
{\em partial} constraints on the ordering, using rule conditions to
finely control the ordering in some cases, e.g., the novel attributes, 
while leaving other cases for the statistics to disambiguate. 
To our knowledge, previous approaches have not addressed informational novelty 
in a conversational system, although we suspect that they could be adapted 
to do so as well.

The hybrid surface NLG module described in this paper is {\em not}
meant to be a general purpose generation package.  Instead, it is
designed to generate utterances in a small domain, such as air travel,
and provides a framework to experiment with the ability to express
additional shades of meaning by varying the word order at run-time.
While our hybrid surface NLG system is not 
linguistically sophisticated as other full-fledged generation packages, the grammar rules are easy
to write, and do not require much  linguistic expertise. 
For this reason, we believe it is more practical than the other full-fledged generation packages.
Our hope is that programmers will be able to implement
NLG in a conversational system without needing to know how to specify detailed linguistic descriptions, as are
usually required by the more sophisticated NLG packages. 
We hope to extend the framework  with some useful facilities as they are
needed by our conversational system.  For example, we hope to add a
facility to pass semantic ``features'' from a parent to a child, and
an interface with a morphological database, which will more properly
deal with phenomena such as agreement and inflection.
Furthermore, we hope to add these extensions without compromising the simplicity of the grammar rule
structure.

\section{Conclusion}

We have presented a system for surface natural language generation that uses
 grammar rules, rule conditions, and statistical information to decide the word
order at run time. 
It takes template fragments given by the programmer, and attempts to 
paste them together in a way that is both consistent with the grammar and optimal with respect
to the scoring function.
We have integrated it into a conversational system for air travel
and have attempted to model the linguistic notion of focus with attribute novelty. 
To our knowledge, we are the first to model informational novelty in a surface generation
system with a combination of grammar rules and statistics, and we are also the first to integrate this
into a practical conversational system. 
Our NLG module is not intended as a general purpose generator, and appears 
adequate for domains of low complexity.
We hope to more extensively use our surface NLG module in our conversational system, 
and we hope that future evaluations will reveal that strategically varying the word
order makes the system talk more like a real person.

\bibliographystyle{acl}
\bibliography{refs}

\end{document}